\documentclass[runningheads]{llncs}
\usepackage{graphicx}
\usepackage{comment}
\usepackage{amsmath,amssymb} 
\usepackage{color}
\usepackage[linesnumbered,ruled,vlined]{algorithm2e}
\usepackage{booktabs,multirow}
\usepackage{subcaption}
\SetKwInput{KwInput}{Input}
\SetKwInput{KwOutput}{Output}

\newcommand{\etal}{\textit{et al. }}

\begin{document}
\pagestyle{headings}
\mainmatter

\title{EasyQuant: Post-training Quantization \\ via Scale Optimization} 


\titlerunning{EasyQuant}
%
\author{Di Wu\inst{1}\and
Qi Tang\inst{2}\and
Yongle Zhao\inst{1}\and
Ming Zhang\inst{1}\and
Ying Fu\inst{3} \and
Debing Zhang\inst{1}
}
\authorrunning{Di Wu et al.}
%
\institute{DeepGlint, Beijing, China \\
\email{\{diwu, yonglezhao, mingzhang, debingzhang\}@deepglint.com} \\
\and OPEN AI LAB, Shanghai, China \\ 
\email{qtang@openailab.com}
\and Beijing Institute of Technology, Beijing, China \\ 
\email{fuying@bit.edu.cn}
}

\maketitle

\begin{abstract}
The 8 bits quantization has been widely applied to accelerate network inference in various deep learning applications. 
There are two kinds of quantization methods, training-based quantization and post-training quantization. Training-based approach suffers from a cumbersome training process, while post-training quantization may lead to unacceptable accuracy drop.
In this paper, we present an efficient and simple post-training method via scale optimization, named EasyQuant (EQ), that could obtain comparable accuracy with the training-based method. Specifically, we first alternately optimize scales of weights and activations for all layers target at convolutional outputs to further obtain the high quantization precision. Then, we lower down bit width to INT7 both for weights and activations, and adopt INT16 intermediate storage and integer Winograd convolution implementation to accelerate inference. Experimental results on various computer vision tasks show that EQ outperforms the TensorRT method and can achieve near INT8 accuracy in 7 bits width post-training.


\end{abstract}

\section{Introduction}
Deep convolutional neural networks (CNN) have made considerable success in various computer vision tasks, including classification, detection, and recognition \cite{squeezenet,mobilenets,vgg,resnet,ssd,insightface}. However, it is not trivial to deploy CNN on computation-constrained devices, due to the huge computing power required by these models. Quantization is an essential technique to reduces CNN models' memory footprint and the amount of computation \cite{han2015deep}.

Low bit representation, e.g. 8 bits width or lower, typically leads to accuracy lost compared with float point 32 (FP32) models  \cite{intel,krishnamoorthi2018quantizing,trt}. Although training-based methods could achieve higher accuracy compared with post-training approaches, they suffer from some drawbacks in real applications. For example, training a quantized neural network is a time-consuming work, and it needs expert experience to guide the whole training process, which significantly influences the success of the work. Besides, in some scenarios, entire training data is not available to deploy the quantization model. 

In this paper, we introduce an efficient and simple post-training quantization method via effectively optimizing the scales of weights and activations.  The proposed scale optimization method is named EasyQuant (EQ). Specifically, we first formulate the quantized convolutional process as an optimization problem target at maximizing the cosine similarity between FP32 and INT8 outputs. This problem is solved by searching weights and activations scales alternately. For entire network optimization, we sequentially optimize scales layer by layer, and greedily obtain the optimal quantization scales for each layer. The scales of weights and activations are jointly optimized in each layer, and their scales of the next layer are optimized based on the quantized results of the previous layers. Besides, we adopt INT16 intermediate storage and integer Winograd algorithm to improve the inference latency on real hardware in context of 7 bits width. Finally, we verify our approach in different bits width settings on common computer vision tasks, including classification, detection, and recognition.

In summary, our main contributions are that 
\begin{itemize}
	\item We present a scale optimization method for the post-training quantization, which alternately searches weights and activations scales target, 
	and can obtain comparable accuracy with the training-based quantization method.
	\item We implement the proposed post-training quantization method to a more efficient INT7 quantization inference framework, which improve the usage efficiency of intermediate INT16 storage.
	\item Extensive experiments on various computer vision tasks demonstrate that our scale optimization approach can achieve effective INT8 post-training quantization and near INT8 precision in the context of 7 bits width without finetuning. Besides, we implement and test the proposed EQ INT7 inference on real ARM platforms.
\end{itemize}
\section{Related Work}
Most works on quantization can be roughly divided into two categories, i.e., training-based quantization and post-training quantization. Training-based quantization usually applies sophisticated designs to train a low bit integer model from scratch or finetune a pre-trained FP32 model \cite{jung2019learning,zhou2017incremental,zhuang2018towards}. Post-training quantization often transforms a pre-trained network from float point to integer range with few data to calibrate applied scales \cite{zhao2019improving,banner2018posttraining,choukroun2019lowbit}. In the following, we review the related work in detail, and discuss the applicability of different quantization methods on real hardware devices.

\subsection{Training-based Quantization}
Early works on training-based quantization often learn a quantization network at a more limited bit width (e.g. under 2 bits) \cite{courbariaux2016binarized,rastegari2016xnor,zhu2016trained}. These methods often suffer from a huge accuracy drop due to the more limited bit width.
Most recent works focus on higher bit width quantization to obtain the similar precision with that from FP32 models \cite{mishra2017wrpn,zhou2017incremental,choi2018pact,jung2019learning}. Mishra~\etal \cite{mishra2017wrpn} proposed wide reduced-precision networks to overcome the accuracy drop by increasing the number of filters and obtain better precision in 4 bits width. Zhou~\etal \cite{zhou2017incremental} proposed to incrementally quantize part of weights of network to reduce the training difficulties involved by quantization in 5 bits width. In \cite{jung2019learning,choi2018pact}, they both adopted the optimizing quantization thresholds by training associated with task loss to guide the training process in 4 bits width. These works all attempt to train less than 8 bits width models from scratch, which are difficult to obtain the similar accuracy with the 8 bits width models. Besides, 
they need specific hardware and software to work, for most of applied devices only support general INT8 quantized models. 
Thus, training-based quantization in lower bits width (4 bits width) is seldom adopted in nowaday industry applications.

Another type of training-based method is quantization aware training (QAT) \cite{krishnamoorthi2018quantizing,jain2019trained}.   \cite{krishnamoorthi2018quantizing} proposed the QAT method as a supplementary approach to regain some lost accuracy induced by INT8 quantization. QAT simulated the quantization noise in conventional training processes and trained the models with normal methods in float point 32 range, which were always used to finetune from an FP32 model \cite{krishnamoorthi2018quantizing}. Jain~\etal \cite{jain2019trained} improved QAT by making the threshold trainable in the regular training process, which could be seen as a training-based scale optimization method. Since these works do not apply sophisticated designs on training process, they can mainly deal with 8 bits width quantization and scarcely consider the effect from the intermediate storage. Besides, 
training a quantized model from scratch has high time complexity and needs the expert experience of both target tasks domain and quantization domain, especially in more complex tasks such as objection detection and face recognition. 

\subsection{Post-training Quantization}
Due to the drawbacks mentioned above, INT8 post-training quantization becomes the major trend in most real quantization applications. Researches on this field include TensorRT (TRT) from Nvidia (Migacz, 2017) \cite{trt} and Tensorflow Lite from Google \cite{krishnamoorthi2018quantizing}.
TRT  \cite{trt}  adopted Kullback-Leibler divergence (KLD) minimization to calibrate quantization thresholds for activations, and utilized the maximum absolute values as thresholds for weights quantization. 
 Tensorflow Lite \cite{krishnamoorthi2018quantizing}
 utilized the maximum absolute values as thresholds for activations, and joined
 a per-channel quantization method with the maximum absolute values as thresholds to quantize the  weight. 
 These two methods quantize the activation and weight scales according to either simple maximum absolute values or statistical characters, which still suffer unaccepted performance drop in some pre-trained networks. 


Later, Yoni Choukroun~\etal \cite{choukroun2019lowbit} improved the quantization method by treating each layer quantization process as a constrained optimization process solved by alternate golden section search. The whole search process is very time-consuming due to the large search space of quantized tensors. Banner~\etal  \cite{banner2018posttraining} optimized the thresholds for activations by theoretically deriving the optimal clipping values. The analytical expressions were based on strict assumptions of activation distribution, which were rarely held in real models.
In this work, we jointly optimize each layers' scales of both weights and activations, and target for reducing the quantization effect of convolutional output. Our method is more robust for various models' situations for not requiring specific hardware and further assumption.

\subsection{Industrial applicability}
One of the significant benefits of quantization techniques is that it could reduce the inference latency on edge devices, which have limited computation power. However, in the literature of quantization, seldom of works discuss the applicability of quantization methods, and it is  non-trivial in real quantization deployment. In order to reduce inference latency in general edge device, e.g. ARM CPU, quantization method provides both quantized weights and activations for convolution operation in inference time. Some approaches, e.g. \cite{zhang2019neural,nayak2019bit,courbariaux2016binarized} which only quantize weights to fixed point, are hard to adopted to accelerate the real inference process. Besides, some methods \cite{rastegari2016xnor,zhao2019improving,choukroun2019lowbit} indeed quantize both weights and activations to fixed point, but they usually need particular hardware or software to facilitate the implementation of quantized inference. This hinders the wide usage of these approaches. Our  method quantizes both weights and activations without the need of specialized hardware. Furthermore, we propose an instruction-level optimization on INT7 quantization inference to accelerate normal INT8 inference which could easily be deployed in general hardware, e.g. the real ARM platforms. 
\section{The Proposed Method}
In this section, 
we first formulate the linear quantization process. Then, the proposed scale optimization method is introduced in detail. Moreover, the designs of the INT7 post-training inference are discussed.

\subsection{Linear Quantization Formulation}
The linear quantization process could be denoted as function $Q(X,S)$, where $X \in \mathbf{R}$ is a tensor and $S$ is a positive real number scale factor. Quantized results $Q(X,S) \in \mathbf{Z}_{b}$, where $\mathbf{Z}_{b}$ is the $b$ bits width integer domain. Linear quantization function $Q(X,S)$ includes three sub-processes, i.e., scaled, round and clipped. The linear quantization formulation of input tensor $X$ and scale factor $S$ could be represented as 
\begin{equation}\label{lq}
Q(X,S) = Clip(Round(X \cdot S)), 
\end{equation}
where $Round$ means that the scaled input tensors are rounded to integers using ceiling rounding, and "$\cdot$" denotes element-wise product. In different implementations of linear quantization, it can adopt different types of rounding (round, ceil, or floor). $Clip$ denotes that elements in the tensor that exceed the ranges of the quantized domain are clipped.

Let us define a quantized $L$-layer neural network as $\{A_{l},W_{l},S_{l}\}_{l=1}^L$. $A_{l}$, $W_{l}$, and $S_{l}$ are the $l$-th layer input activation, weight, and quantization scale factors in FP32 range, respectively.  
Specifically, quantization scale factors ($S_{l}$) contain two parts. The scale number for the input activations in the $l$-th layer is denoted as $S_{l}^a$. The scale number for the weights in the $l$-th layer is denoted as $S_{l}^w$. $S_{l}^a$ is a non-negative real number applied for each elements in feature maps. $S_{l}^w$ is a non-negative real number for input weights. For the convenience of discussion, we illustrate the  proposed method in per-layer quantization scheme.  
For per-channel quantization scheme, each filter in $W_{l}$ should have independent scale number. Besides, we denote the $l$-th output feature map of the pre-trained FP32 model as $O_{l}$ and its corresponding quantized inference output feature map as $\hat{O}_{l}$.

Therefore, the whole linear quantization forward convolution and dequant operation in the $l$-th layer can be described as 
\begin{equation}
\hat{O}_{l} = \frac{Q(A_{l},S_{l}^a) \ast Q(W_{l},S_{l}^w)}{S_{l}^a \cdot S_{l}^w}.
\label{eqlq}
\end{equation}
where $\ast$ denotes convolution operation.
The original output of the $l$-th layer can  be expressed as 
\begin{equation}
O_{l} = A_{l} \ast W_{l}.
\label{eqo}
\end{equation}

From Equation \eqref{eqlq}, it can be seen that the scale factors actually control the thresholds clipped in quantization process, which  affects the cosine similarity of convolutional results between original output feature map ($O_{l}$) and quantization inference feature map ($\hat{O}_{l}$) to a great extent. Therefore, our target function focuses on optimizing the scale factors for both weights ($S_{l}^w$) and activations ($S_{l}^a$) and improving the similarity between $O_{l}$ and $\hat{O}_{l}$.

\subsection{Scale Optimization}
\begin{figure}[h]
\begin{center}
   \includegraphics[width=1\linewidth]{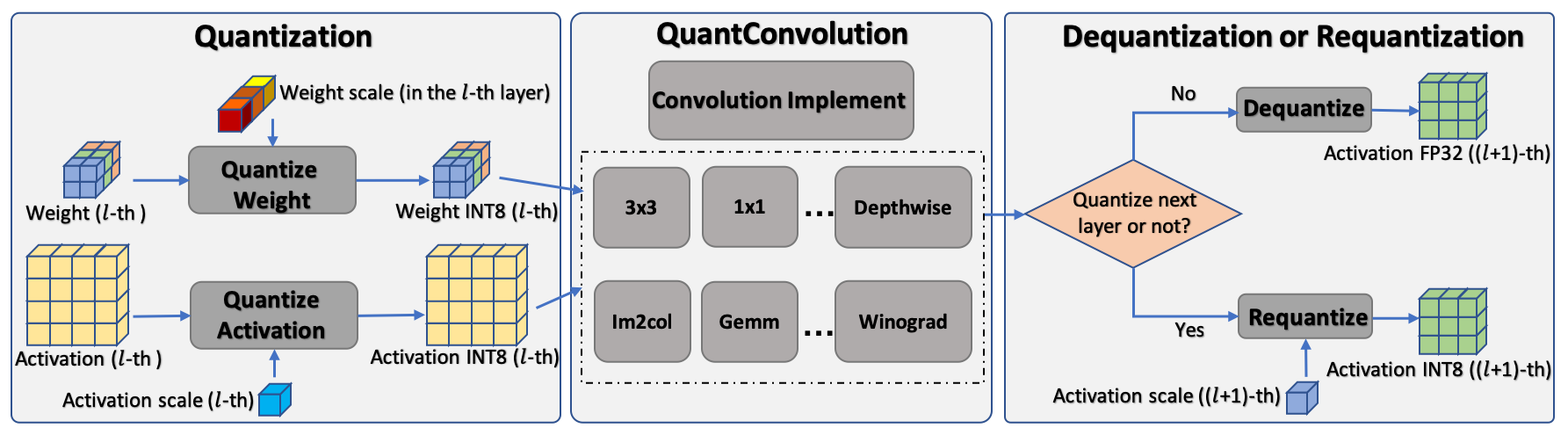}
\end{center}
   \caption{The quantization process of a convolutional layer. The whole process contains three parts: quantization of input activations and weights, convolutional operation implementation and requant/dequant operation.}
\label{linear quantization}
\end{figure}

Quantization process of a neural network model could be divided into each layer, where weights and activations are quantized respectively and prepared for convolutional operation. The quantization of a convolutional layer is shown in Figure \ref{linear quantization}.

The overall Information is passed and transformed from the first layer of the neural network to the end. It introduces inevitable noise into the final outputs of the neural network. Most post-training works \cite{trt,ncnn,MNN} use the Kullback-Leibler divergence (KLD) method to calculate scale factors of activation for each layer. Typically, they use around 1000 calibration data to approximate the input activation distribution of each layer \cite{trt}. For the scales of weights in each layer, they usually use the absolute maximum as thresholds to determine scales because of the bigger weights values always dominate the results. They separately optimize the scale for each activation and do not optimize weight scales, which easily results in error accumulation. It also ignores the fact that the optimization of similarity between original and quantized distribution, which can not guarantee the promotion of the similarity between original and quantized convolutional outputs. Furthermore, these approaches mainly design INT8 quantization on activations and weights, and need INT32 bits to save all the intermediate results in the CNN inference process. This takes more computational time and limits their
application fields. 

To solve this problem, we present a simple but efficient scale optimization method, which jointly optimize both scales of activation and weights targeting at the loss on the similarity between original and quantized convolutional outputs. Moreover, we also propose INT7 quantization inference to further accelerate traditional INT8 quantization and deploy it on real ARM platforms. 

\subsubsection{Optimal Scale for Each Layer}
According to Equation \eqref{eqlq},  we optimize scale factors $\{S_{l}\}_{l=1}^L$ for a pre-trained convolutional neural network $\{A_{l},W_{l}\}_{l=1}^L$, where activations $\{A_{l}\}_{l=1}^L$ are generated from a given calibration dataset $D$ with $N$ samples. It is much less than the common training dataset. 
In the $l$-th layer of the neural network, our approach can be expressed to maximize  output feature map cosine similarity
\begin{equation}
\begin{aligned}
\max_{S_{l}} \quad & \frac{1}{N} \sum_{i = 1}^{N} \cos{(O_{l}^i,\hat{O}_{l}^i)},\\
\textrm{s.t.} \quad & S_{l} \in \mathbf{R^{+}}.\\
\label{eqtarget}
\end{aligned}
\end{equation} 
where $S_{l}$ scales the activations and weights to the fixed bits width range. Generally, the larger the scale is, the more high value elements will be saturated to the maximum of the quantized domain. While the smaller the scale is, the more low value elements will be rounded to zeros.

We adopt alternating optimization method here to solve this problem in two folds. First, $S_{l}^a$ is fixed, and solve $S_{l}^w$ for weight scales adjustment. Second, $S_{l}^w$ is fixed, and solve $S_{l}^a$ to finetune activations scales. $S_{l}^w$ and $S_{l}^a$ are alternately optimized until $\cos{(O_{l}^i,\hat{O}_{l}^i)}$ converges or excesses the time limitation. 
Here, for fast convergence, $S_{l}^w$ and $S_{l}^a$ are initialized in terms of the maximum of weigths or activations respectively. 
For search space of $S_{l}^w$ and $S_{l}^a$, we linearly divide interval of $\left[ \alpha S_{l},\beta S_{l} \right]$ into $n$ candidate options and conduct a simple search strategy on them. In experiments, hyper-parameters $\alpha, \beta$ and $n$ are robust for various tasks with $\alpha=0.5$, $\beta=2$ and $n=100$. 
More advanced search method could be applied to search the candidate scales, while we find that in experiments, simple search strategy is more robust for the irregular fluctuation of target function. Therefore, these strategy with reasonable initialization is applied in our optimization process to solve the problem. When optimizing weight scales in per-channel quantization scheme, where $S_{l}^w$ is a collection of $c$ (the number of filters in this layer) dimension, we can adjust each kernel's independent scale in parallel in one search process.

\subsubsection{Optimal Scale for the Whole Network}
In the previous section, we present how to optimize one layer activation scale $S_{l}^a$ and weight scale $S_{l}^w$ with optimal scale in a layer. We apply our approach layer by layer sequentially for a whole convolutional neural network.
\begin{algorithm}
\DontPrintSemicolon
  \KwInput{model weights set $\{W_{l}\}_{l=1}^L$, model original input activation set and output set$\{A_{l},O_{l}\}_{l=1}^L$ generated by calibration $D$}
  \KwOutput{Optimal scales for $L$ layers $\{S_{l}\}_{l=1}^L$} 
  \KwData{Calibration dataset $D$ with $N$ samples} 
  Initialize $\{S_{l}\}_{l=1}^L$ for $L$ layers\\
        \While{Convergence or excess time limitation}
            {
            \For{$l = 1:L$ }    
                { 
                    \For{$S_{lk}^w$ in interval $\left[ \alpha S_{l}^w,\beta S_{l}^w \right]$}
                        {
                            record ${S_{lk}^{w*}}$ with
                            maximum $\frac{1}{N} \sum_{i \in \mathbb{D}} \cos{(O_{l}^i,\hat{O}_{l}^i)}$ 
                        }
                }
            Fix $\{S_{l}^{w*}\}_{l=1}^L$ optimize $\{S_{l}^{a}\}_{l=1}^L$ and update $A_{l+1}$ with $\hat{A}_{l+1}$\\
            \For{$l = 1:L$ }
                {
                    \For{$S_{lk}^a$ in interval $\left[ \alpha S_{l}^a,\beta S_{l}^a
                    \right]$}
                        {
                            record ${S_{lk}^{a*}}$ with
                            maximum $\frac{1}{N} \sum_{i \in \mathbb{D}} \cos{(O_{l}^i,\hat{O}_{l}^i)}$
                        }
                }
            }
  \Return{Optimal searched scales for $L$ layers $\{S_{l}\}_{l=1}^L$}
\caption{Scale optimization for the whole convolutional neural network}
\label{alpp}
\end{algorithm}

For each layer, input activation $\hat{A}_{l}$ is obtained from the current model, where all former layers are quantized and optimized. The output feature map $O_{l}^i$ is collected from the original model without quantization. The benifits of adopting this greedy strategy is that (a) dividing the optimization of entire networks to subproblems helps reduce the huge search space of optimization, and (b) the optimization of current layer $O_{l}^i$ is taking accumulated noise from all former layers into consideration. Above all, for an $L$ layers convolutional neural network, we adopt our proposed approach  sequentially to obtain the optimal scales $\{S_{l}\}_{l=1}^L$. The proposed scales optimization for post-training layers in the integral model quantization is summarized in Algorithm \ref{alpp}. 

\subsection{INT7 Post-training Inference}
Here, we implement our efficient designs on INT7 post-training inference. Detailed explanations are also discussed here to give more insight on the significance of 7 bits width.

\begin{figure}[h]
\begin{center}
   \includegraphics[width=1\linewidth]{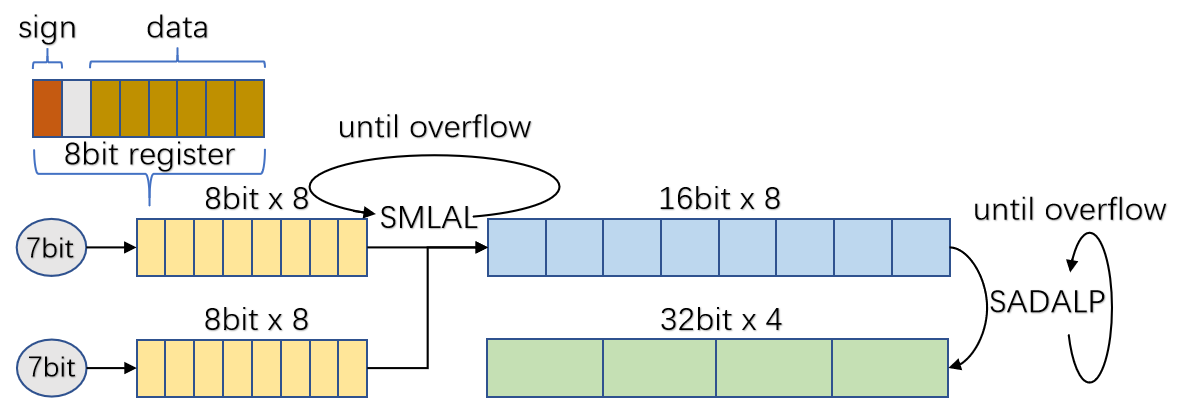}
\end{center}
   \caption{Proposed INT7 inference data flow in ARM architecture. Two main NEON instructions applied in implementation of convolution are SMLAL and SADALP. In case of overflow, 7 bits inference could do 8 times SMLAL instruction (2 times for 8 bits) safely before adding intermediate 16 bits results to 32 bits register by SADALP.}
\label{int7inference}
\end{figure}

Quantization relies heavily on the characteristic of hardware in order to take advantage of low bits inference. 
In regular convolutional calculations, there are many matrix multiply-add operations which could be implemented by Signed Vector Multiply-Add Long instruction (SMLAL) and Signed Add and Accumulate Long Pairwise instruction (SADALP) in ARM NEON instruction sets. SMLAL instruction multiply and add 8-bit elements to produces 16-bit results while SADALP instruction adds two adjacent 16-bit results into 32-bit accumulators \cite{googleint16}. Using these two instructions, we could efficiently implement convolutional process on ARM architecture. The data flow mentioned above is described in Figure \ref{int7inference}.

In INT8 quantization, the safe solution is to use 32-bit register to store the intermediate variable. 
Nevertheless, before ARM V8.2-A architecture in Cortex-A processor, there is no instruction to store the multiplication result of two 8-bit register numbers into a 32-bit register. Therefore, the general solution uses SMLAL first to multiply-add 8-bit elements and produces 16-bit results, and then uses SADALP to add two adjacent 16-bit results into 32-bit accumulators \cite{googleint16}. For regular convolution operation, 8 bits (8 bits signed integer) inference could only make $\left \lfloor{\frac{2^{15}-1}{(2^{7}-1)^2}}\right \rfloor\ = 2$ times SMLAL operations without any overflow which is ineffective \cite{ncnn}. Our proposed 7 bits (7 bits signed integer) inference could do $\left \lfloor{\frac{2^{15}-1}{(2^6-1)^2}}\right \rfloor = 8$ times SMLAL without overflow which is more effective compared with INT8 inference. It can make more SMLAL operations before adding them to 32-bit accumulators compared with 8 bits inference. In general, INT7 post-training method could make more use of CPU efficiency \cite{googleint16} , which is important in industrial application.
\section{Experimental Results}
In this section, 
we first describe the setting applied in our following experiments, including the method compared in our experiments, quantization implementation details and hyper-parameters in our EQ. Then, we  conduct experiments on different bit-width settings, including most common INT8 bits width, proposed INT7 bits width, and less than 7 bits width. 
\subsection{Settings}
Our method belongs to post-training approach due to no requirement of retraining. To make fair comparison, we compare our EQ with the post-training approach TRT \cite{trt,trtweb,tf} \footnote{In the post-training methods, the TRT approach is widely adopted for optimizing activation scales in real industrial deployment, for most inference frameworks utilize the TRT method to optimize activation scales \cite{trt,ncnn,MNN}. In particular, the TRT method compared in our experiments refer to current version, which include per-channel quantization}. 1000 samples are used for calibration data on TRT method in all experiments. 

In our experiments, we alternately optimize weights and activations for one round and use 50 random samples to optimize the scales targeting at convolutional outputs. For the number of samples, we have tested our method that more samples give almost same results in 8 bit width and can provide a little better results in 7 bit width. For example, there are about 0.1\% $\sim$ 0.15\% gain of MobileNet v1 on ImageNet2012 with INT7 quantization, by using 1000 samples instead of 50. Considering the much time consuming of 1000 samples and limited computation resource we have, we use 50 samples on all the experiments, which have outperformed TRT method with 1000 samples. And we search 100 values uniformly distributed between $\left[ 0.5 S_{l},2 S_{l} \right]$ for each scales in one layer.

According to implementations, all methods are tested on the open-source quantization framework NCNN \cite{ncnn}. Given $n$ bit quantization, we constrain the absolute maximum values under $2^n-1$, which saturates the values beyond the threshold. In TRT approach, for activation scales, 
1000 samples randomly drawn from the respective training set are used to generate the activations scales $\{S_{l}\}_{l=1}^L$.

\subsection{INT8 Post-training Quantization}
8 bits post-training quantization is adopted widely, when deploying the convolutional model in a general edge device or specific server optimized engine \cite{trt,tf,ncnn,intel}. Here, we test our method in context of 8 bits quantization. concretely, we compare our EQ with widely used TRT \cite{trt,trtweb,tf} and quantization aware training (QAT) \cite{krishnamoorthi2018quantizing} approach.  We re-implement TRT quantization scheme with NCNN framework based on their open resources \cite{ncnn,trt,trtweb}. For QAT comparison, we directly adopt the results in \cite{krishnamoorthi2018quantizing} for reference. 

\begin{table}[h]
\caption{Top 1 classification accuracy (\%) on ImageNet2012 validation dataset for different convolutional models in context of both INT8 and INT7 post-training quantization. Bold results show the better results between our EQ and TRT.} 
\centering 
\begin{tabular}{ccccccc} %
\toprule
\multirow{2}{*}{
\parbox[c]{.15\linewidth}{\centering Models}} &
\multirow{2}{*}{
\parbox[c]{.15\linewidth}{\centering FP32}}
  & \multicolumn{2}{c}{INT8} &&
\multicolumn{2}{c}{INT7} \\ 
\cmidrule{3-4} \cmidrule{6-7}

 & & {\centering TRT} & {EQ} && {TRT} & {EQ}  \\
\midrule
SqueezeNetV1.1 \cite{squeezenet}& 56.56  & 56.24 & \textbf{56.28} && 54.88 & \textbf{56.08} \\
MobileNetV1 \cite{mobilenets} & 69.33  & 68.74 & \textbf{68.84} && 66.97 & \textbf{68.26} \\
VGG16 \cite{vgg} & 70.97  & 70.95 & \textbf{70.97} && 70.92 & \textbf{70.96} \\
ResNet50 \cite{resnet} & 75.20  & 75.04 & \textbf{75.13} && 72.78 & \textbf{75.04} \\
\bottomrule
\label{imagenetINT87}
\end{tabular}
\end{table}

We verify the effectiveness of our method among different tasks, including image classification (ImageNet2012)\footnote{Most previous quantization methods \cite{courbariaux2016binarized,rastegari2016xnor,banner2018posttraining,zhou2017incremental,zhao2019improving,choukroun2019lowbit,krishnamoorthi2018quantizing} only test the performance on image classification tasks, e.g. on the ImageNet2012 dataset. }, object detection (VOC2007), and face recognition (seven stardard). Performances are evaluated on ImageNet 2012 validation dataset \cite{russakovsky2015imagenet}, Pascal VOC object detection 2007 test dataset \cite{everingham2010pascal} and seven common face recognition datasets \cite{huang2008labeled,moschoglou2017agedb,zheng2017cross,sengupta2016frontal,zheng2018cross,cao2018vggface2}. For model architectures, more Computationally efficient backbone MobileNet V1 \cite{mobilenets} are chosen among all the tasks. Meanwhile, other classical models e.g. SqueezeNetV1.1 \cite{squeezenet}, ResNet50 \cite{resnet} and VGG16 \cite{vgg} are also tested in our experiments.

\begin{table}[h]
\caption{Object detection on VOC2007 task for SSD \cite{ssd} models with backbone SqueezeNet and MobileNet V1. Mean average precision (mAP) is evaluated in FP32, TRT and our EQ (both in INT8 and INT7 post-training quantization).} 
\centering 
\begin{tabular}{ccccccc} %
\toprule
\multirow{2}{*}{
\parbox[c]{.15\linewidth}{\centering Models}} &
\multirow{2}{*}{
\parbox[c]{.15\linewidth}{\centering FP32}}
  & \multicolumn{2}{c}{INT8} &&
\multicolumn{2}{c}{INT7} \\ 
\cmidrule{3-4} \cmidrule{6-7}

 & & {\centering TRT} & {EQ} && {TRT} & {EQ}  \\
\midrule
SqueezeNet-SSD& 62.00  & 61.45 & \textbf{62.05} && 60.01 & \textbf{61.62} \\
MobileNet-SSD & 72.04  & 69.79 & \textbf{71.39} && 63.88 & \textbf{68.79} \\
\bottomrule
\label{vocINT87}
\end{tabular}
\end{table}

For 8 bits quantization, the results of the classification on ImageNet2012 \cite{deng2009imagenet}, detection on VOC2007, and seven standard face recognition tasks are shown in Tables \ref{imagenetINT87}, \ref{vocINT87} and \ref{faceINT87}, respectively. It can be seen that our EQ outperforms  TRT method on per-channel INT8 quantization across all three kinds of tested tasks and all the convolutional neural network architectures. For example, for MobileNet V1 backbone models, EQ could further gain 0.1 \%, 1.6\% and average 0.37 \% precision compared with TRT respectively. It implies, in some complex task, e.g. object detection, EQ could reduce more precision loss compared with TRT method. 

\begin{table}[h]
\caption{Verification performance (\%) for InsightFace \cite{insightface} model MobileFaceNet on seven most common validation datasets. Comparisons are on TRT and EQ with INT8 and INT7 quantization.} 
\centering 
\begin{tabular}{ccccccc} %
\toprule
\multirow{2}{*}{
\parbox[c]{.15\linewidth}{\centering Test dataset}} &
\multirow{2}{*}{
\parbox[c]{.15\linewidth}{\centering FP32}}
  & \multicolumn{2}{c}{INT8} &&
\multicolumn{2}{c}{INT7} \\ 
\cmidrule{3-4} \cmidrule{6-7}

 & & {\centering TRT} & {EQ} && {TRT} & {EQ}  \\
\midrule
lfw& 99.45  & 99.36 & \textbf{99.48} && 99.28 & \textbf{99.36} \\
agedb\_30 & 95.78  & 95.23 & \textbf{95.38} && 95.03 & \textbf{95.73} \\
calfw & 95.05  & 94.76 & \textbf{94.88} && \textbf{94.75} & 94.68 \\
cfp\_ff & 99.50  & 99.50 & \textbf{99.61} && 99.44 & \textbf{99.60} \\
cfp\_fp & 89.77  & 89.17 & \textbf{90.04} && 88.47 & \textbf{89.87} \\
cplfw & 86.45  & 85.58 & \textbf{86.03} && 85.91 & \textbf{86.76} \\
vgg2\_fp & 90.64  & 89.70 & \textbf{90.50} && 89.64 & \textbf{90.44} \\
\bottomrule
\label{faceINT87}
\end{tabular}
\end{table}

We also compare our method with more complex QAT approach in 8 bit width. The results of MobileNetV1 and ResNet50 models on ImageNet classification are shown in Table \ref{qat}.  The results show that EQ could get competitive precision compared with more complex QAT method in context of INT8 quantization. It is noticeable that for ResNet50 EQ could even outperform QAT method.

\begin{table}[h!]
\caption{Top 1 classification accuracy (\%) on ImageNet2012 validation dataset for MobileNetV1 and ResNet50 with EQ INT8 method and QAT approach. Bold results show the better results between EQ and QAT.} 
\centering 
\begin{tabular}{cccccc} %
\toprule
\multirow{2}{*}{
\parbox[c]{.15\linewidth}{\centering Methods}}
  & \multicolumn{2}{c}{MobileNetV1} &&
\multicolumn{2}{c}{ResNet50} \\ 
\cmidrule{2-3} \cmidrule{5-6}

 & {FP32} & {INT8} && {FP32} & {INT8}  \\
\midrule
EQ & 69.33  & 68.84 && 75.20 & \textbf{75.13} \\
QAT \cite{krishnamoorthi2018quantizing} & 70.90  & \textbf{70.70} && 75.20 & 75.00 \\
\bottomrule
\label{qat}
\end{tabular}
\end{table}

\subsection{INT7 Post-training Quantization}
Both requiring activations and weights constrained in 7 bits keep less information from the original model and will add more quantization errors into models. Therefore, it require more advanced quantization scheme in INT7 post-training quantization. In this section, we test our method with TRT in context of 7 bits width. Besides, we also verify the latency of our proposed INT7 quantization inference on real hardware. The experiments are conducted on the same three tasks and respective models with INT8 post-training quantization. 

\begin{table}[h!]
\begin{center}
\caption{The latency (ms) performance on RK3399, whose inside is a 1.5 GHz 64-bit Quad-core ARM Cortex-A53. \#k means k threads.}
\begin{tabular}{*{5}{c}}
\hline
Models & TRT-INT8(\#1) & EQ-INT7(\#1) & TRT-INT8(\#4) & EQ-INT7(\#4)\\
\hline\hline
SqueezeNetV1.1 & 180 & \textbf{120} & 66 & \textbf{44}\\
MobileNetV1 & 234 & \textbf{189}& 65 & \textbf{57}\\
VGG16 & 3326 & \textbf{2873}& 1423 & \textbf{1252}\\
ResNet50 & 1264 & \textbf{993}& 415 & \textbf{300}\\
\hline
\label{speedA53}
\end{tabular}
\end{center}
\end{table}

The results of these experiments are shown in Tables \ref{imagenetINT87}, \ref{vocINT87}, and  \ref{faceINT87}. In these INT7 quantization experiments, TRT method suffers a sharp accuracy drop compared with FP32 model. For the classification on ImageNet2012  and detection on VOC2007, the accuracy lost is even more significant in some models, e.g. MobileNet V1 (2.36\% and 8.16\% in ImageNet2012 and VOC2007, respectively). Our EQ shows substantial superiority that it can obtain much better precision on all the models across three tasks (largest gap occurred in MobileNet V1 1.06\% and 3.25\% in ImageNet2012 and VOC2007 respectively). Our EQ performs much better compared with TRT method in INT7 quantization, and our proposed method could still achieve near FP32 accuracy in 7 bits width.

 As we point out in section 3.3, INT7 quantization reduces 2 bits storage for activations and weights, and give solid supports for Int16 intermediate storage. Its inference could be more efficient compared with INT8 inference. In our EQ INT7 quantization, the summation of 8 multiplication results could be saved directly into INT16 without overflow. Using INT16 registers is not only much faster than using INT32 registers, but also efficiently reduces the amount of memory access which is also very important. Concretely, about 20\%-33\% computational cost could be saved on various ARM platforms. We test the INT7 post-training inference latency based on RK3399 \footnote{RK3399 is a low power, high performance processor based on ARM architecture}. The results are provided in Tables \ref{speedA53} and \ref{speedA72}. It can be seen that our proposed INT7 inference scheme has less latency on general edge devices.
 
\begin{table}[h!]
\begin{center}
\caption{The latency (ms) performance on RK3399, which inside is a 1.8 GHz 64-bit Dual-core ARM Cortex-A72. \#k means k threads.}
\begin{tabular}{*{5}{c}}
\hline
Models & TRT-INT8(\#1) & EQ-INT7(\#1) & TRT-INT8(\#2) & EQ-INT7(\#2)\\
\hline\hline
SqueezeNetV1.1 & 79 & \textbf{57} & 54 & \textbf{37}\\
MobileNetV1 & 105 & \textbf{84} & 56 & \textbf{46}\\
VGG16 & 1659 & \textbf{1385} & 1034 & \textbf{849}\\
ResNet50 & 559 & \textbf{463} & 338 & \textbf{262}\\
\hline
\label{speedA72}
\end{tabular}
\end{center}
\end{table}

\subsection{Comparison on Less than 7 Bits Width}
\begin{figure}[h!]
    \centering
    \begin{subfigure}{0.32\textwidth}
        \centering
        \includegraphics[width=1\linewidth]{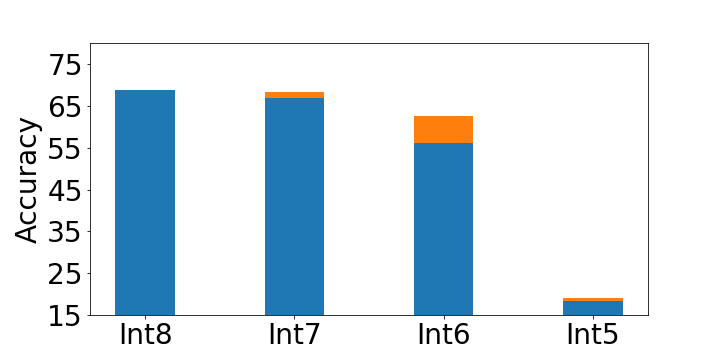}
        \caption{}
        \label{less7bitconbinec}
    \end{subfigure}
    \begin{subfigure}{0.32\textwidth}
        \centering
        \includegraphics[width=1\linewidth]{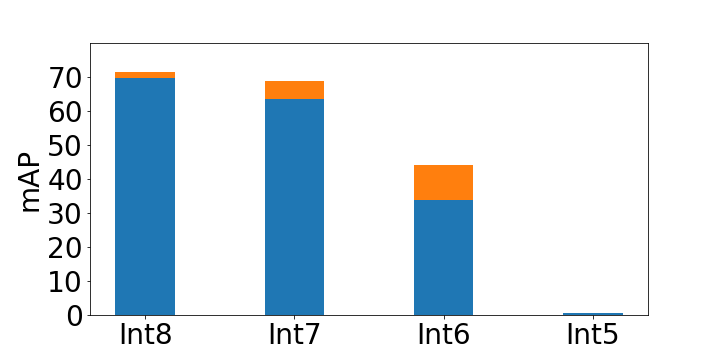}
        \caption{}
        \label{less7bitconbined}
    \end{subfigure}
    \begin{subfigure}{0.32\textwidth}
        \centering
        \includegraphics[width=1\linewidth]{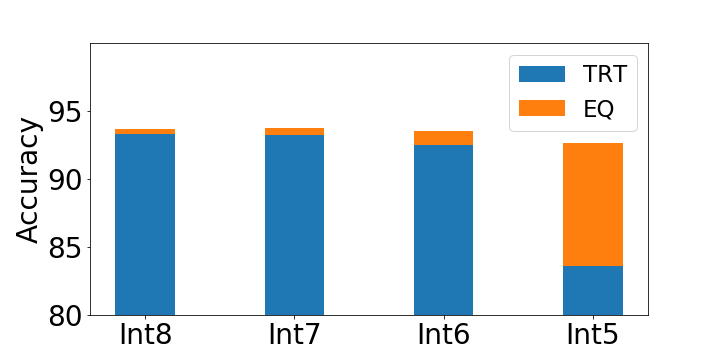}
        \caption{}
        \label{less7bitconbinef}
    \end{subfigure}
    \caption{Comparison between TRT and EQ in less than 7 bits width on classification, detection and Face recognition. (a) Top 1 classification accuracy (\%) for MobileNet V1 models tested on ImageNet2012 validation dataset of different bit widths. (b) mAP (\%) for SSD models with MobileNet V1 backbone on VOC2007 test dataset of different bit widths. (c) Mean verification accuracy (\%) for InsightFace model MobileFaceNet on seven test datasets of different bit widths.}
    \label{less7bitconbine}
\end{figure}

\begin{figure}[h!]
    \centering
    \begin{tabular}{ccccc}
    TRT \quad &
    \begin{subfigure}{0.2\textwidth}
        \centering
        \includegraphics[width=1\linewidth]{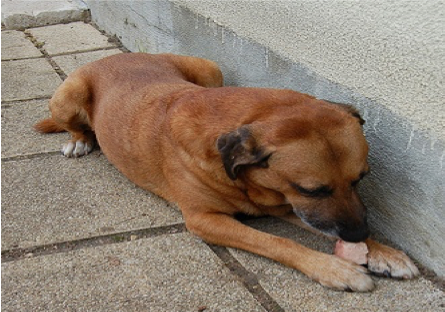}
    \end{subfigure} &
    \begin{subfigure}{0.2\textwidth}
        \centering
        \includegraphics[width=1\linewidth]{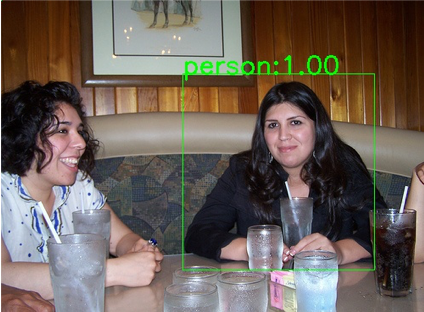}
    \end{subfigure} &
    \begin{subfigure}{0.2\textwidth}
        \centering
        \includegraphics[width=1\linewidth]{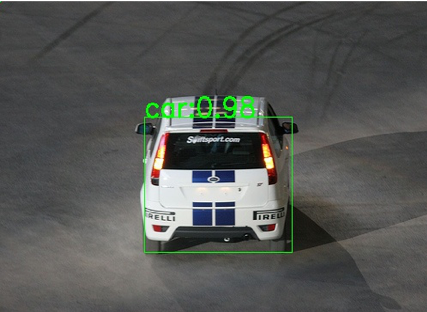}
    \end{subfigure} &
    \begin{subfigure}{0.2\textwidth}
        \centering
        \includegraphics[width=1\linewidth]{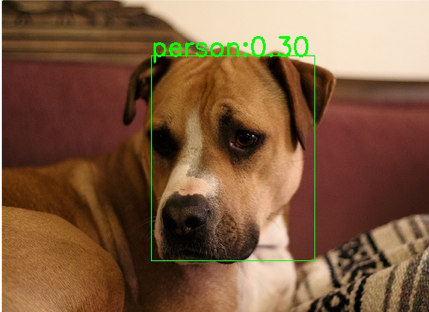}
    \end{subfigure}\\
    EQ \quad&
    \begin{subfigure}{0.2\textwidth}
        \centering
        \includegraphics[width=1\linewidth]{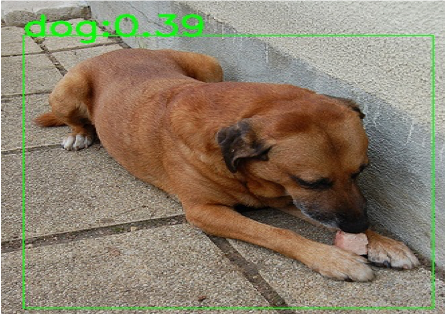}
        \caption{}
    \end{subfigure} &
    \begin{subfigure}{0.2\textwidth}
        \centering
        \includegraphics[width=1\linewidth]{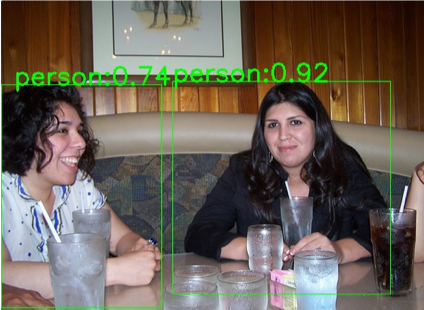}
        \caption{}
    \end{subfigure} &
    \begin{subfigure}{0.2\textwidth}
        \centering
        \includegraphics[width=1\linewidth]{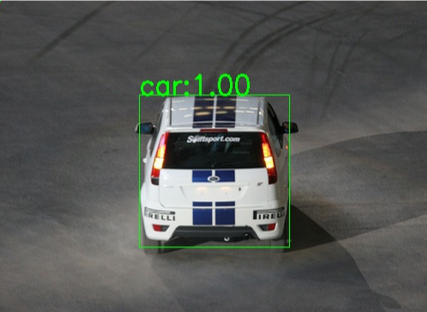}
        \caption{}
    \end{subfigure} &
    \begin{subfigure}{0.2\textwidth}
        \centering
        \includegraphics[width=1\linewidth]{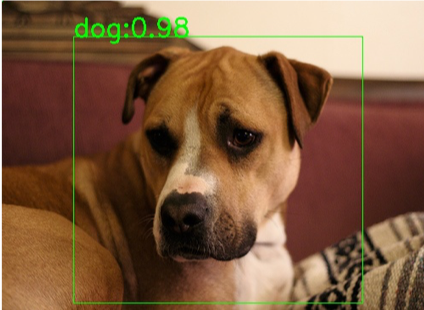}
        \caption{}
    \end{subfigure}\\
    \end{tabular}
    \caption{Visualizations (Mobilenet V1 SSD model) of different types of detection errors in 6 bits quantization. (a) False negative of detection in single object scenario. (b) False negative of detection in multiple objects scenario. (c) Inaccurate bounding box regression. (d) Wrong type of detection.}
    \label{deterrors}
\end{figure}

We also conduct experiments with less than 7 bits width on classification, detection and face recognition. 
Taking MobileNet V1 based model as an example, the results are shown in Figure \ref{less7bitconbine}. 
For some tasks (classification on ImageNet2012 and detection on VOC2007), when the bits width is constrained to less than 7 bits, the precision dropped sharply on these tasks, as shown in Figure \ref{less7bitconbine}(a) and \ref{less7bitconbine}(b). In particular, on the VOC2007 tasks, the mean average precision (mAP) drops sharply when the bits width is constrained to less than 7 bits.
The visualizations of 6 bits width MobileNet SSD model quantized by TRT method and our EQ approach are shown in Figure \ref{deterrors}. It can be seen that our EQ is still better than TRT method.

However, In face recognition task, It can be seen that even lower than 7 bits, our EQ still outperforms TRT method in all bits width we tested, especially, EQ with 6 bits width on insight face model still keeps near FP32 precision on seven face recognition datasets, as shown in Figure \ref{less7bitconbine}(c). In real application of post-training quantization, one could choose the appropriate bits width to balance the acceleration of deployment and lost precision introduced by quantization.

\section{Conclusion}
In this paper, we present a scale optimization based method to boost post-training quantization both from the perspective of preserved quantization precision and deployment latency. Our proposed INT7 quantization inference does not rely on any specific framework and could be applied to any linear post-training scheme to both increase the inference speed and accuracy. It benefits real industrial INT8 post-training quantization without complex quantization aware finetuning. 
Experiments show that our proposed method can obtain better precision of quantized models in various tasks and convolutional architectures. By designing Int16 intermediate storage and integer Winograd algorithm, we can further improve the inference speed with less precision decreasing compared with TRT methods on real hardware platform.
%
%
\bibliography{eccv2020submission}
\end{document}